%% file: anonymous-submission-latex-2025.tex
\documentclass[letterpaper]{article} % DO NOT CHANGE THIS
\usepackage{aaai25}  % DO NOT CHANGE THIS
\usepackage{times}  % DO NOT CHANGE THIS
\usepackage{helvet}  % DO NOT CHANGE THIS
\usepackage{courier}  % DO NOT CHANGE THIS
\usepackage[hyphens]{url}  % DO NOT CHANGE THIS
\usepackage{graphicx} % DO NOT CHANGE THIS
\usepackage{natbib}  % DO NOT CHANGE THIS AND DO NOT ADD ANY OPTIONS TO IT
\usepackage{caption} % DO NOT CHANGE THIS AND DO NOT ADD ANY OPTIONS TO IT
\usepackage{algorithm}
\usepackage{algorithmic}
\usepackage{amsmath}
\usepackage{amsfonts}
\usepackage{booktabs}
\usepackage{multirow}

\usepackage{newfloat}
\usepackage{listings}
\DeclareCaptionStyle{ruled}{labelfont=normalfont,labelsep=colon,strut=off} % DO NOT CHANGE THIS
\floatstyle{ruled}
\newfloat{listing}{tb}{lst}{}
\floatname{listing}{Listing}
\title{Safe Planner: Empowering Safety Awareness in Large Pre-Trained Models for Robot Task Planning}
\author{
   Siyuan Li\textsuperscript{\rm 1}, Zhe Ma\textsuperscript{\rm 2}, Feifan Liu\textsuperscript{\rm 1}, Jiani Lu\textsuperscript{\rm 1}, Qinqin Xiao\textsuperscript{\rm 1}, Kewu Sun\textsuperscript{\rm 1}, Lingfei Cui\textsuperscript{\rm 3}, Xirui Yang\textsuperscript{\rm 2},\\
    Peng Liu\textsuperscript{\rm 1}, Xun Wang\textsuperscript{\rm 2}\\
    \textsuperscript{\rm 1}Harbin Institute of Technology
    \\
    \textsuperscript{\rm 2}Intelligent Science \& Technology Academy Limited of CASIC
    \\
    \textsuperscript{\rm 3}Institute of Computer Application Technology, Norinco Group
}
\affiliations{
}

\usepackage{bibentry}
\begin{document}

\maketitle

\begin{abstract}
Robot task planning is an important problem for autonomous robots in long-horizon challenging tasks.
As large pre-trained models have demonstrated superior planning ability, recent research investigates utilizing large models to achieve autonomous planning for robots in diverse tasks. 
However, since the large models are pre-trained with Internet data and lack the knowledge of real task scenes, large models as planners may make unsafe decisions that hurt the robots and the surrounding environments. 
To solve this challenge, we propose a novel Safe Planner framework, which empowers safety awareness in large pre-trained models to accomplish safe and executable planning. 
In this framework, we develop a safety prediction module to guide the high-level large model planner, and this safety module trained in a simulator can be effectively transferred to real-world tasks. 
The proposed Safe Planner framework is evaluated on both simulated environments and real robots. The experiment results demonstrate that Safe Planner not only achieves state-of-the-art task success rates, but also substantially improves safety during task execution. The experiment videos are shown in \url{https://sites.google.com/view/safeplanner}.

\end{abstract}


\section{Introduction}

Robot task planning is a challenging temporal planning problem \cite{kaelbling2013integrated}, aiming at obtaining a subtask sequence to accomplish a long-horizon target task, which is of great importance in the autonomous robot domain \cite{guo2023recent}. Classical robot task planning methods are mostly based on search or STRIPS \cite{fikes1993strips}. However, these classical methods require much domain-specific knowledge and can hardly be applied to dynamic real-world environments. Following task-planning researchers pay attention to learning-based methods, e.g., reinforcement learning and imitation learning \cite{ceola2019robot, mcdonald2022guided, jiang2019task}. Although these learning-based methods have shown promising results, they usually overfit the training scenes and cannot support natural languages as target task descriptions. Recently as large pre-trained models have demonstrated superior planning abilities in general settings, more and more researchers \cite{guan2023leveraging, zhao2024large, song2023llm} try to employ large pre-trained models to achieve autonomous robot task planning with languages as instructions. 

% 概述现有大模型planning的方法怎么做的，存在什么问题（safety），举个例子（图）说明safety问题

\begin{figure}[t!]
\centering
\includegraphics[width=0.48\textwidth]{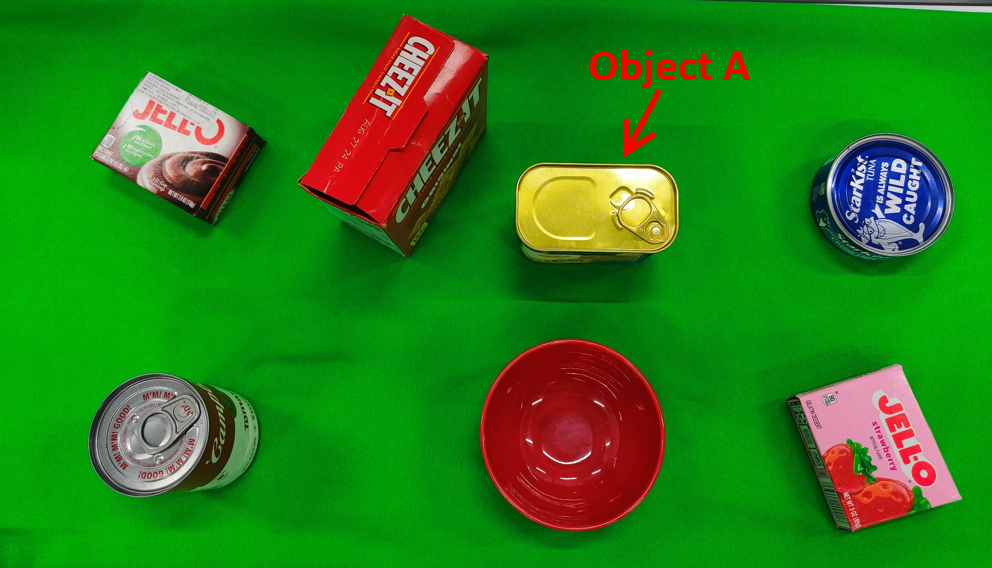} 
\caption{An illustrative example of safe planning.}
\label{fig1}
\end{figure}

% 补充一个实物场景的对比图
Recent robot task planning methods based on large pre-trained models mainly utilize vision-language models (VLMs) and large language models (LLMs) to understand the task scenes and decompose the challenging long-horizon tasks \cite{driess2023palm, hu2023look, huang2023inner}. 
These models are pre-trained with large-scale data, and thus are endowed with great generalization abilities. However,  as lacking real-world training data,
it is quite difficult for these models to accurately understand the task scenes, which may lead to unsafe and unexecutable planning results, causing damage to the robots and the real-world environments.  
For example, in a moving-object task shown in Figure \ref{fig1}, a bunch of objects has been put on one table, and the task is to move object $A$ to a target position. 
The pre-trained models without real scene knowledge may output a direct planning solution, e.g., picking object $A$, and then placing it at the target place. However, as object $A$ is surrounded by other objects, directly moving $A$ may damage the nearby objects or cause collisions. 
Therefore, only using the pre-trained models is not enough, and
a smarter planning model with safety awareness is needed, which is expected to move the objects nearby object $A$ to other places before moving the target object $A$. 

% 讲我们的方法是如何解决以上问题的，提出什么样的方法，达到什么样的实验效果。
To achieve safe planning in general robotics tasks, we propose a novel Safe Planner framework, which empowers safety awareness in large model based robot task planning. 
This framework is comprised of two levels: the high-level planner, and the low-level executors (skills). 
The high-level planner takes natural language task instruction and visual observations as inputs, and outputs the next skill to execute. For a structured interface between the two levels, we use the planning domain definition language (PDDL) \cite{fox2003pddl2} to define skills. 
As only using the large models pre-trained with Internet data as planners suffer from the unsafe problem, we develop a safety prediction module, which can predict the safety of executing the low-level skills. Then, the safety prediction results are leveraged to guide the planning process of the large pre-trained models, incentivize the reasoning ability of the large models, and thus promote safe planning in general robot tasks. Note that in real-world robot tasks, most unsafety is caused by collisions. Therefore in this work, the safety measure is defined with collision numbers, and the proposed framework can be easily adapted with a broader safety definition. 

In the experiments, we compare the proposed framework with state-of-the-art large model planning methods in both the simulated environments and the real robot manipulation tasks. The experiment results demonstrate that Safe Planner not only achieves better task success rates, but also substantially improves safety during task execution. 
 Besides, The safety prediction module trained with the simulated data can be effectively transferred to the real robot settings. 
To further investigate the safety module, we conduct thorough ablation studies on its design.

\section{Preliminaries and Problem Statement}
This section first briefly describes the preliminary knowledge for this work, including VLMs and PDDL, and then presents the problem statement.

\subsection{Vision-Language Models}
First, we introduce language models, which try to model the probability $p(W)$ of a text $W=\{w_0, w_1, w_2, ..., w_n\}$, a sequence of strings $w$. The probability $p(W)$ is generally modeled by the chain rule, $p(W)=\Pi_{j=0}^np(w_j|w_{<j})$, so that each successive string can be predicted from the previous. Recent breakthroughs induced by the Attention mechanism \cite{vaswani2017attention} and Transformer architecture \cite{devlin2019bert} have enabled the efficient scaling of LLMs \cite{achiam2023gpt,touvron2023llama,chowdhery2022palm}, which have demonstrated increasingly large capacity and generalization ability. For real tasks grounded in the physical world, only the language modality is not enough, and VLMs augment LLMs with visual inputs to achieve the understanding of both images and languages. The visual inputs are processed with a vision encoder, such as ResNet \cite{alayrac2022flamingo} and vision transformers (ViT) \cite{driess2023palm,chen2023pali}. In this work, we employ an LLM, GPT-4, and a VLM, GPT-4v to accomplish the proposed safe robot planning framework for physically grounded tasks.

\subsection{Planning Domain Definition Language}

The Planning Domain Definition Language (PDDL) \cite{fox2003pddl2} serves as a standardized encoding of planning problems. A PDDL planning problem is represented by two parts: a domain and a problem. Next, we describe these two terms informally and refer interested readers to comprehensive guides \cite{geffner2022concise}. 
\begin{itemize}
\item A PDDL \textit{domain} defines the ``universal'' aspects of a problem, i.e., the elements in all specific problems such as object types, predicates, and operators. Types and predicates are used to describe the world state.
For example, in a \textit{House-Keeping} domain, the predicates may include ``\textit{in(X, Y): Is object X in container Y?}'' and ``\textit{holding(X): Is the robot holding object X?}''. An operator specifies a change to the state and is typically structured into three parts: preconditions, postconditions, and parameters. The preconditions determine whether the operator is applicable, and the postconditions define the effect of executing the operator. In the House-Keeping domain, the operators may include \textit{pick(X)}, \textit{place(X, Y)}, and \textit{Open(X)}, and these operators can be regarded as robot skills.
\item A PDDL problem specifies a list of objects, an initial state $s_0$, and a goal state $s_g$. Both $s_0$ and $s_g$ are composed of a set of predicates. The types in the domain are used to describe the objects in the problem.
\end{itemize}
As PDDL predicates and operators have a clear structure, PDDL has been shown an effective interface in the LLM-based planning paradigm \cite{silver2024generalized,liu2023llm+}, which has alleviated the LLM hallucination problem and improved the generalization ability. Beyond that, since PDDL types, predicates, objects, and operators often include human-readable names like the ones above, PDDL boosts the ability of LLM to solve embodied planning problems. 

\begin{figure*}[htbp]
\centering
\includegraphics[width=0.98\textwidth]{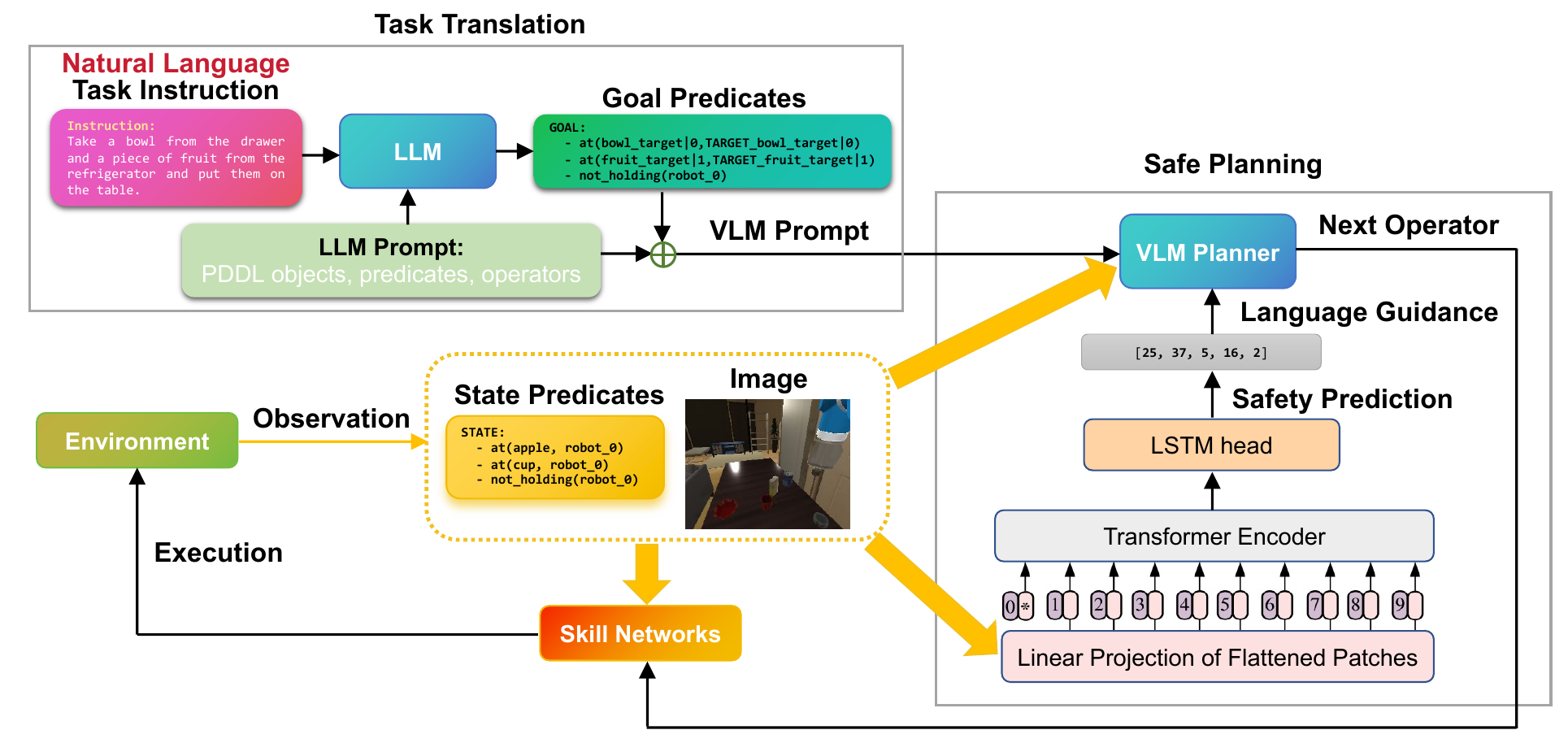} 
\caption{The Safe Planner framework. First, our framework translates the natural language task instruction into the PDDL goal predicates. Then, the goal predicates are combined with the PDDL domain as the prompt for the VLM planner. Note that in the Safe Planner framework, the VLM planner not only takes the observations as inputs, but also considers the safety of the operators, and selects the next operator to execute.}
\label{frame}
\end{figure*}

\subsection{Problem Statement}
Assume that a set of planning tasks are in the same PDDL domain, the goals of these planning tasks are different, which are given in the form of natural language. This setting is common in the real world, e.g., a home service robot needs to do multiple household tasks, where these tasks are in the same house scenario, but have different goals, and these goals are described with natural languages. 
 The planning framework aims to obtain a plan for each task, which can lead to both task completion and high safety. Here safety means that when the robot executes the plan, it will not damage the environment or itself.

\section{Approach}

Recently rich literature has tried to employ pre-trained LLMs and VLMs to accomplish comprehensive planning in complex long-horizon robotics tasks. In robotics tasks, the robot needs to interact with the physical world, but the models pre-trained with Internet data can hardly understand the real-world environments. Therefore, utilizing the large pre-trained models as a robotic task planner may hurt the robot or the environment. To address this problem, we propose a novel Safe Planner framework shown in Figure \ref{frame}. In this framework, the natural language task description is translated into the PDDL goals with an LLM. Taking the PDDL instruction, the current observation, and the safety prediction from the safety module as inputs, a VLM task planner outputs a selected skill in the form of PDDL operators, and the operator is executed with low-level skills in the environment. After the skill execution completes or exceeds a preset timestep, the VLM planner replans in a closed-loop way. 
The following subsections elaborate on task translation and safe planning shown in the gray boxes in Figure \ref{frame}.

\subsection{Task Translation}

Considering a robotic household scenario, human users usually describe the target tasks with natural languages, but for the robot, natural languages are not executable. Therefore, first of all, we need to translate the natural language task instruction to the language executable by robots. As PDDL is a well-structured planning language with high executability, we develop a PDDL-based task translator, as shown in the left-up box in Figure \ref{frame}.

The inputs and outputs of the task translator are both languages. As the pre-trained LLMs have shown substantial language processing and reasoning abilities, we employ a pre-trained LLM as the main part of the task translator. To inspire the LLM to translate the task description into PDDL goals, we have designed a prompt template, which includes the PDDL domain information, the PDDL grammar, the translation requirement, and a successful translation example. Since the prompt template is quite long, we present it in the Appendix.
The natural language task description, prompted by the designed template, is processed by an LLM to generate the corresponding PDDL goals. We utilize GPT-4 as the LLM in the task translator, and other pre-trained LLMs are applicable in the Safe Planner framework as well. 

\begin{figure*}[t!]
\centering
\includegraphics[width=0.98\textwidth]{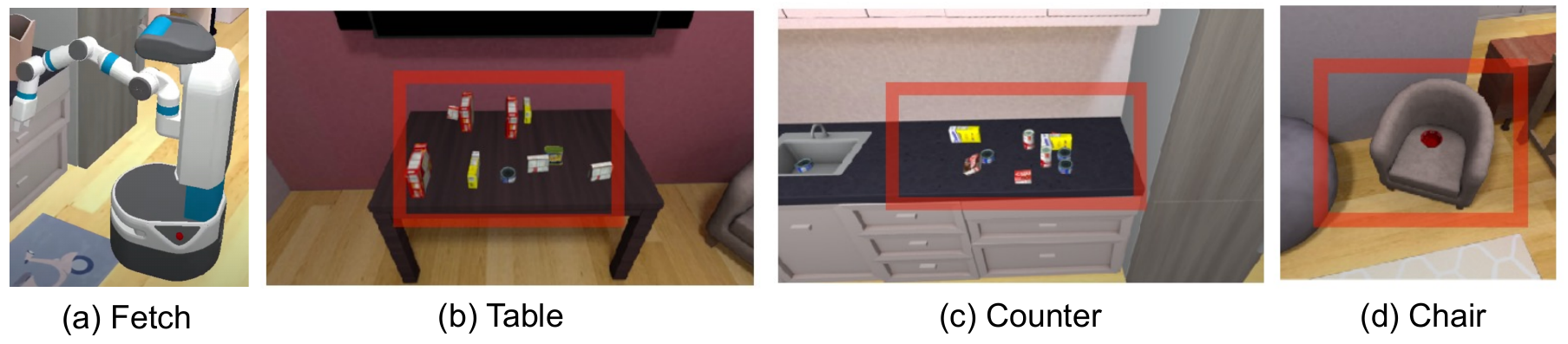} 
\caption{The robot and target task scenes in the simulated experiments.}
\label{fig3}
\end{figure*}

\subsection{Safe Planning}

% 对于safety的定义，收集数据集(这里应该补充safety定义)，(模型，训练)，怎么影响VLM planning

Robot safety has long been a central issue for autonomous robots \cite{tzafestas2013introduction}. As a broad concept, robot safety involves a lot of factors: collision with environments, collision with humans, velocity limits, force/torque limits, etc. Since unsafe actions with limit violations can be prohibited by a properly set action range, in this work we focus on robot safety concerning collisions, which is of great importance especially in dynamic environments.  To achieve safe planning in long-horizon complex tasks, we propose a novel safety prediction module, which can predict the safety of executing skills based on the current observation. Then, the safety prediction can guide the VLM planner to accomplish safe and executable planning, as shown in the right part of Figure \ref{frame}. Next, we elaborate on dataset collection, safety prediction module, and guidance to the VLM planner.

\textbf{Dataset Collection:} To diversify the training data, we randomize the initial scene of the tasks during data collection, i.e., the objects and the robot positions vary among episodes. In these diverse task scenes, the pre-trained low-level skills (corresponding to the PDDL operators) are executed to generate a trajectory set $D$. Then, we calculate the risk level of these trajectories as follows\footnote{$N$ denotes the number of objects in the robot available area.}.
\begin{equation}
Risk(i, o) = C(robot,o^-) + \sum_{j,k}C(o_j, o_k)|_{j,k\in 1...N, j\neq k},
\label{eq1}
\end{equation}
where $Risk(i, o)$ denotes the risk of executing the $i$-th skill to manipulate the object $o$, and $o^-$ denotes the objects except $o$ in the current scene. $C$ represents the collision number, which is calculated empirically with the collected trajectories in the simulators. Considering the influence of the robot on the scene $C(robot,o^-)$ and the induced collisions among the $N$ objects $\sum_{j,k}C(o_j, o_k)|_{j,k\in 1...N, j\neq k}$, the risk level $Risk(i, o)$ is defined as Equation \eqref{eq1}, which provides the labels for training the safety prediction module.
% % 这里之后应该需要改
% Specifically, we divide the trajectory set $D$ with the median of $Risk(i, o)$ into two parts: the trajectories with lower $Risk$ are labeled as safe ($\hat{y}=1$), and the ones with higher $Risk$ are labeled as unsafe ($\hat{y}=0$).
% Note that the median as the threshold induces the balance of the training dataset. 
% With these labeled data, we formulate safety prediction as a binary classification problem and train the safety module. 

\textbf{Safety Prediction Module:} 
% 明确输入、输出、损失函数等
The safety prediction module is a multi-head neural network $f_{\theta}(obs, o; i)|i\in 1...I$, where the $i$-th head predicts the safety of executing the $i$-th skill in the skill set of size $I$, $o$ is the object to manipulate and $obs$ is the initial image observation in the trajectory of skill execution. To enable a comprehensive description of the object $o$, we represent $o$ with images of the object from five different views, so the input of the safety prediction module is a set of images. 
In the safety prediction module, a large-sized ViT encoder \cite{wu2020visual} pre-trained with the ImageNet dataset \cite{deng2009imagenet} is employed to extract effective representations from the input images, and then the representations are processed with a learnable LSTM network \cite{hochreiter1997long} to predict the risk level $Risk(i, o)$ in Equation \eqref{eq1}. Here we use the LSTM network since the inputs are treated as a sequence of images. 
Regarding the safety prediction problem as a regression problem, the LSTM is optimized with the mean squared error (MSE) loss, and the pre-trained ViT encoder is frozen.

% As the inputs of the safety prediction module are all images, we utilize a standard vision model, ResNet18 \cite{he2016deep}, as the backbone of the safety network, and its loss function $L_\theta$ is the cross entropy loss.
% \begin{equation}
% L_\theta = -\mathbb{E}_{D}[\hat{y}\cdot p +(1-\hat{y})\cdot log(1-p)],
% \end{equation}
% where $p=f_{\theta}(obs, o; i)$ denotes the safety probability, and the safety network $f_\theta$ is activated by the Sigmoid function. 

\textbf{Guidance to VLM:} To achieve safe planning, the VLM planner requires guidance from the safety prediction module.
As a thorough safety prediction, we formulate the guidance with a matrix of size $I\times N$, where the rows correspond to the pre-trained skills, and the columns correspond to the objects in the available area of the robot. The safety matrix can be obtained by inferring the safety module with the $N$ objects. 
The next question is how to guide the VLM planner with this safety matrix. 
As the exact values in this matrix are difficult to understand by VLMs, we transform the numeric values in the matrix into rankings in the form of natural languages. This transformation from 
the safety matrix to the ranking sentences relies on the PDDL operator names and object names.
An example feedback of the safety module to the VLM planner is shown as follows:
% as follows:

% Then, each element in the safety matrix is given to the VLM planner with sentences pointing out the skill and the object, e.g., 

\textit{ The safest operator is to pick the bowl. The second safest operator is to pick the apple.$\cdots$}

Besides the safety guidance, the inputs of the VLM planner include the task information from the task translation module as the prompt for VLM, and the current observation of state predicates and images. The details about the VLM prompt are provided in the Appendix. With these inputs, the VLM planner outputs the next PDDL operator to execute. Then, the pre-trained low-level skill corresponding to the selected operator is executed. If the skill execution satisfies the success criteria or exceeds a certain timestep, the VLM replans with the renewed observation and safety matrix.
In this work, we use GPT-4v as the VLM planner and utilize the PPO algorithm \cite{schulman2017proximal} to pre-train low-level skills. Note that other VLMs and skill pretaining algorithms are also applicable.

\input{tables/sim_colli.tex}

\input{tables/sim_sr.tex}

\section{Related Work}
% 前面总写一句话
This work is mostly related to large model planning and robot safety. Next, we give a brief literature review about these two directions, and discuss the relationship between our work and previous literature.

% LLM planning
\textbf{Planning with Large Models.} 
Recently, robot task planning with large pre-trained models has drawn much attention from embodied artificial intelligence researchers \cite{driess2023palm, song2023llm, guan2023leveraging, wang2024llm}. In most related works, the large pre-trained models serve as a high-level planner, which instructs low-level executors \cite{brohan2023can, hu2023look, singh2023progprompt, huang2023inner}. To alleviate the hallucination issue in LLM planning, existing works propose to use structured interfaces to regularize the outputs of LLM, such as PDDL \cite{liu2023llm+, silver2024generalized, pallagani2022plansformer}. However, these methods lack the real scene grounding, and thus may output plans that hurt the robot or environment. To address this problem, \citet{huang2023grounded} propose Grounded Decoding, which guided the LLM planning with a hand-crafted safety score. 
The following works propose to employ VLMs to enhance the scene understanding abilities and improve planning safety \cite{hu2023look, chen2023open}. In contrast to the previous works, the proposed Safe Planner framework explicitly learns a safety prediction module, which can guide the VLM planner to achieve safe planning in complex long-horizon tasks.

% safety in robotics, 需要再查点资料，robotics safety分成几个方面，现有工作都是怎么做的
\textbf{Safety in Robotics.} Safety has long been a critical issue for autonomous robots, especially in the human-robot interaction setting \cite{tzafestas2013introduction, vicentini2021collaborative}. Robot safety involves multiple perspectives, e.g., mechanical limitations, collisions, etc. This work considers robot safety from the collision avoidance view. An early related work builds high-level symbolic representations that capture physical interactions to avoid collisions \cite{mojtahedzadeh2015support}. The following work develops an object-level scene understanding approach, SafePicking, to improve manipulation safety \cite{wada2022safepicking}. A recent work proposes to build a relation graph of the whole scene to promote robot safety \cite{li2024broadcasting}. These related works mainly address the safety problem with a small model, which limits their applicability to general settings. To the best of our knowledge, the proposed framework is the first to induce a learnable safety prediction module into large model planning, which injects safety awareness into general robotics planning with large models.

\section{Experiments}

% 把exp的outline列出来，出实验结果直接往里面填
In this section, we conduct experiments to answer the following questions: (1) Can the Safe Planner framework achieve state-of-the-art performance in challenging tasks?  (2) Is the proposed framework applicable to real robots? (3) How is the planning result with safety awareness compared to that without safety awareness? (4) How can the safety module design influence the planning performance? Next, we briefly review the baseline methods and then demonstrate the experiment results.

\subsection{Baselines}
We compare Safe Planner with state-of-the-art methods in the large model planning domain, and all the comparison methods are listed as follows. For a fair comparison, in all the methods, the LLMs are implemented with GPT-4, and the VLMs are implemented with GPT-4v.
\begin{itemize}
% 写成“方法简称（加上引用）”： “描述的形式”
% 描述问题需要体现该方法和我们方法的联系和区别
% 简称需要和后面实验结果中对应
\item Safe Planner: Empowering safety awareness in large models with a learnable safety prediction module.
    \item Safe Planner w/o SM: Removing the safety module (SM) in the proposed framework, which can be regarded as an ablation study.
    \item SayCan \cite{brohan2023can}: LLM planner with value functions as affordances and languages as interfaces.
    % \item Inner Monologue \cite{huang2023inner}: Augmenting SayCan with grounded closed-loop language feedbacks.
    % \item NLMap \cite{chen2023open}: Enhancing the semantic understanding ability of SayCan with open-vocabulary queryable scene representations.
    \item PROGPROMPT \cite{singh2023progprompt}: Prompting LLMs to output Pythonic task plans.  
\end{itemize}

\subsection{Results in Simulated Environments}
% 先写实验设置，然后列一张实验结果表，然后写一段分析文字
% 实验场景： habitat, robot, task（）, objects, 放三个场景的图
\textbf{Task Settings:} The simulated experiments are conducted in the Habitat 2.0 environment \cite{szot2021habitat}, where a mobile manipulator (Fetch robot) equipped with two RGBD cameras mounted on its head and arm is instructed to do housework. The experiments in this subsection involve three target task scenes, as shown in Figure \ref{fig3}(b)-(d). Due to the tight spaces, the \textit{Chair} and \textit{Counter} scenes are more difficult than the \textit{Table} scene.
The long-horizon challenging tasks in the experiments are composed of both navigation subtasks and manipulation subtasks, i.e., the robot needs to first navigate to the target task scene, and then manipulate the target objects as instructed by the natural languages, e.g., ``\textit{Move the apple on the table to the chair}''. The detailed natural language task instructions are listed in the Appendix.
As a thorough evaluation of Safe Planner, we design two task modes with different difficulty levels in each target task scene: in the \textit{``easy''} mode, there are totally $3$ objects in the manipulation areas denoted with the red rectangles in Figure \ref{fig3}; in the \textit{``hard''} mode, there are no less than $5$ objects in the manipulation areas. More objects lead to complex obstacle geometry and collisions easily happening, and thus are reckoned as \textit{hard}.

% 对比的measure: task success rates, collision, 如何统计的，多少次取平均
\textbf{Metrics:}  We use two metrics to compare the performance of these planning methods: task success rates to evaluate executability, and the total collision numbers in one trajectory as calculated by Equation \eqref{eq1} to evaluate safety. The average results of $100$ runs are shown in Tables \ref{sim_colli} and \ref{sim_sr}.

% 分析性文字：对比w/o SM体现SM的作用，w/o SM和PROGPROMPT对比：体现pddl作用。另外三个baseline都是saycan一个系列的，一起讨论一下
\textbf{Analysis:} The experiment results demonstrate that the proposed Safe Planner framework can effectively reduce the collision numbers during task execution and output planning results with higher safety. The improvement of safety (reducing collisions) is at a little sacrifice of success rates, since with safety awareness, the skill sequences output by large models are longer than those without safety awareness, as the Safe Planner first moves the obstacles away to ensure the safe execution. An example planning result is shown in the next subsection, and the longer skill sequence leads to a lower success rate in the whole task due to the chaining rule.  

Comparing the last three columns in Tables \ref{sim_colli} and \ref{sim_sr}, we find that without the safety prediction module, the collision number of the proposed framework is nearly the same as the baseline methods. However, the success rates of Safe Planner w/o SM are larger, which indicates the PDDL interface of Safe Planner is more effective than the natural language interface of SayCan and the Pythonic interface of PROGPROMPT. 

\begin{figure}[htbp]
\centering
\includegraphics[width=0.45\textwidth]{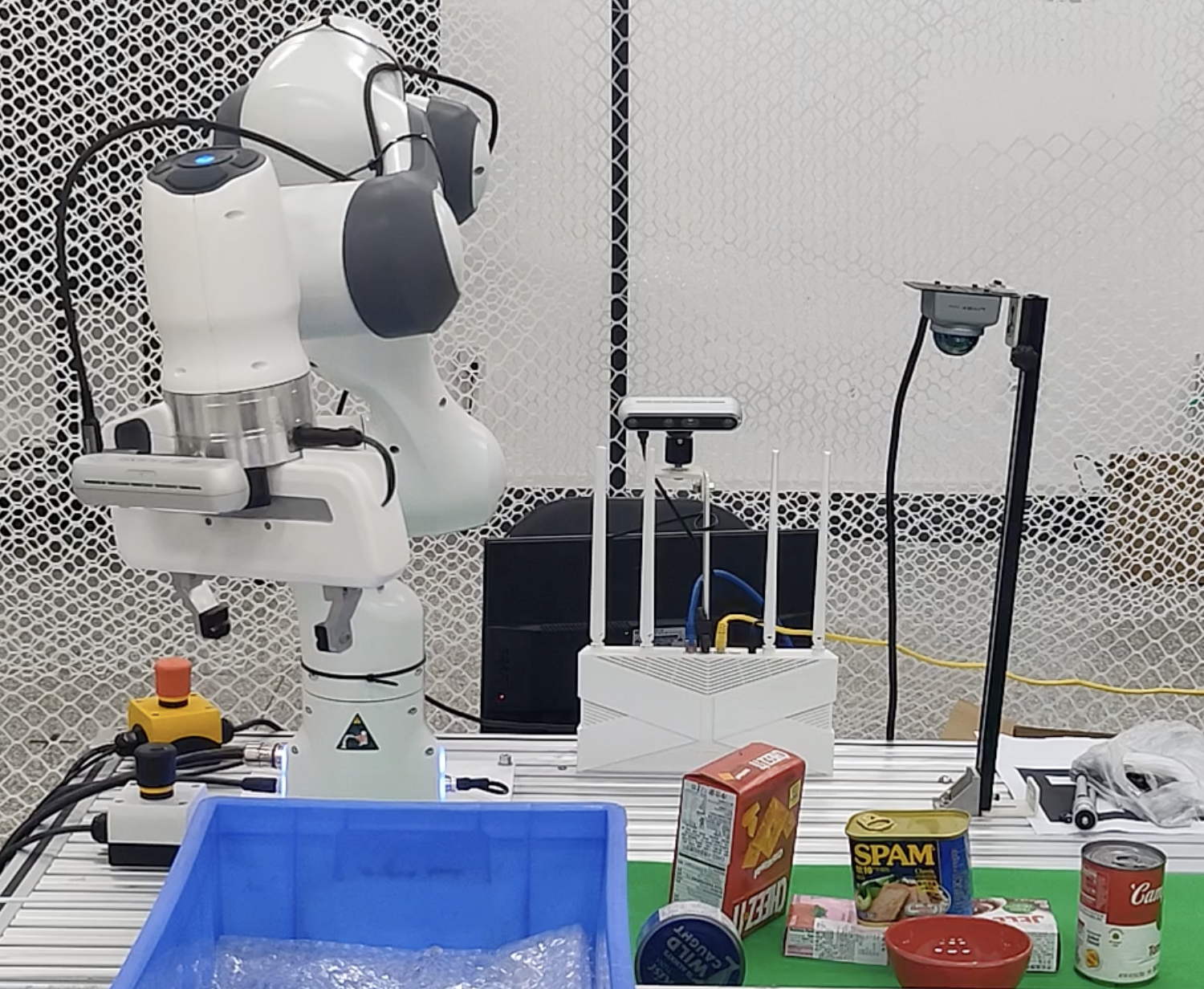} 
\caption{The task scene for real robot experiments.}
\label{robot}
\end{figure}

\subsection{Results on Real Robots}
% 先写实验设置，然后列一张实验结果表，然后写一段分析文字
\textbf{Task Settings:} The real-world experiments are conducted on a fixed-based 7-dof Franka Panda robot arm, as shown in Figure \ref{robot}. As the collision data on the real robots is difficult to obtain, the safety prediction module trained with the simulated data is directly transferred to the real-world setting in a zero-shot manner without fine-tuning. 
The tasks in this subsection are about robot arm manipulation. Specifically, according to a natural language task instruction, the robot arm needs to pick the target object from a cluttered table, and then place this object at the target place.
To achieve a thorough evaluation, the table is set with different levels of clutter: in the \textit{easy} mode, there are $3$ objects on the table, so accomplishing the Pick-Place task without hurting the surrounding objects is relatively easy. In the setting with \textit{medium} difficulty, there are $5$ objects on the table, and in the \textit{hard} one, there are $7$ objects\footnote{More detailed real robot task settings with photos are listed in the Appendix.}. 
Note that in the real robot experiments and the simulated experiments, we use different types of robots, so transferring the manipulation skills from sim to real is quite challenging. Therefore, for the real robot arm, the low-level manipulation skills are implemented by hand-crafted scripts with object pose estimation.

\textbf{Metrics:} In the real robot experiments, we also evaluate the performance from two perspectives: task success rates and safety. As the collision number in the real-world setting is not easy to calculate, we measure the safety by examining the final state after task execution. If the robot collides with one object unrelated to the task, which causes the object position changed in the final state, we add $1$ to the \textit{collision} metric. If the collision is more severe, which causes the turning over of the object, we add $2$ to the \textit{collision} metric. The collisions with all the task-unrelated objects are summed together as the \textit{``collisions''} metric. The results averaged over $10$ runs are shown in Tables \ref{real_colli} and \ref{real_sr}.

\input{tables/real_colli.tex}
\input{tables/real.tex}

\textbf{Analysis:} Table \ref{real_colli} indicates that the safety module (SM) can significantly reduce the collision numbers and improve safety during manipulation in all the task settings. As the task success is measured by the final state of the target objects (no matter with the surrounding objects), in the easy and medium settings, with and without SM both can achieve a $100\%$ success rate. However, the success of without SM is at the cost of collisions.  
The success rates in the easy and medium settings are a little higher than those in the simulated environment, since the low-level skills on the real robot with hand-crafted scripts are more accurate than the skills trained with the PPO method in the simulator.
In the hard task setting with more objects on the table, the planner without SM can hardly accomplish the manipulation tasks.

\begin{figure}[htbp]
\centering
\includegraphics[width=0.5\textwidth]{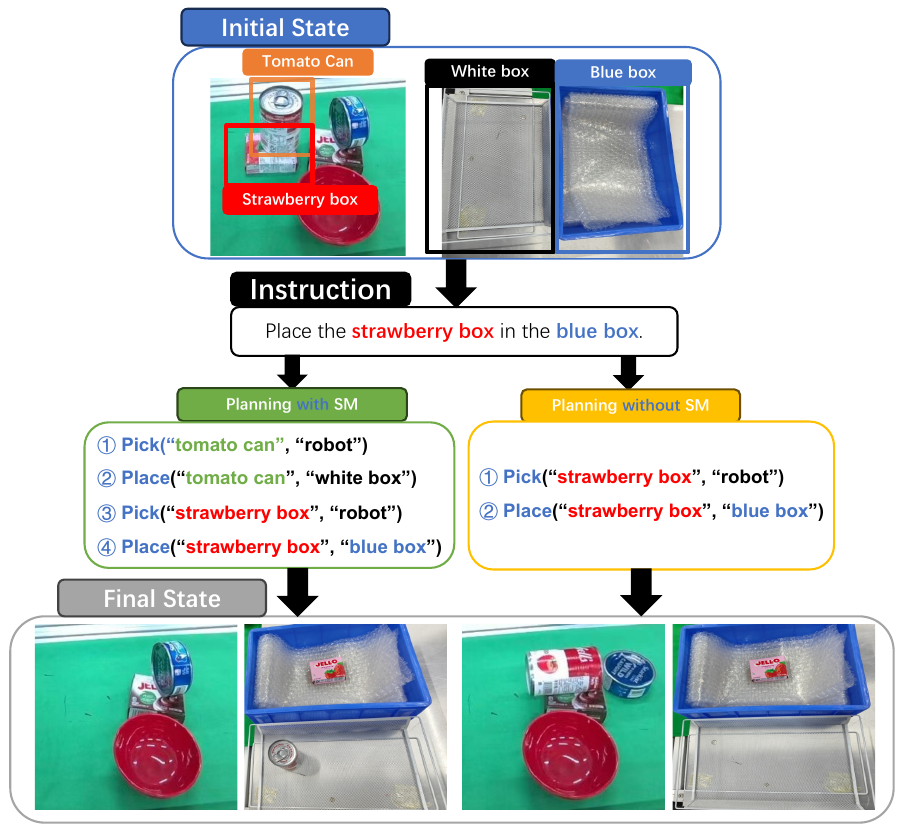} 
\caption{Comparison of the skill sequences planned by the model with SM (left) and without SM (right).}
\label{example}
\end{figure}
% 这部分可以补充一个任务上的planning results对比，用了safety model是什么样的subtask sequence, 不用的是什么样子的
\textbf{Skill Sequence Comparison:}
% 前面讲一段话，比较的是什么(为什么做这个比较)，然后放图，最后做点分析（skill sequence长度，w/SM规划出把障碍物挪走， final state区别）
Diving into why Safe Planner has such great performance, we showcase the planning results on an example task (the task with \textit{medium} difficulty). As shown in Figure \ref{example}, the objects are cluttered on the table at the initial state, and the instruction is to put the strawberry box under the tomato can into the blue box. Without safety awareness, the planner directly picks and places the target object (the strawberry box), which causes collisions with the tomato can and the blue can, as shown in the right part. In contrast, guided by the SM, the planner first moves the tomato can away, and then manipulates the strawberry box, which keeps the objects surrounding the strawberry box safe, as shown in the left part of Figure \ref{example}. The videos for more tasks are shown in \url{https://sites.google.com/view/safeplanner}.

\subsection{Ablation Studies on Safety Module Design}
% 对safety module的模型(CNN or transformer)和是否pretrain做ablation study,并分析
% 插一个图和一个表：图体现收敛的速度，表体现regression loss 和collision之间的关系（这是在哪个task上做的？）

\begin{figure}[htbp]
\centering
\includegraphics[width=0.48\textwidth]{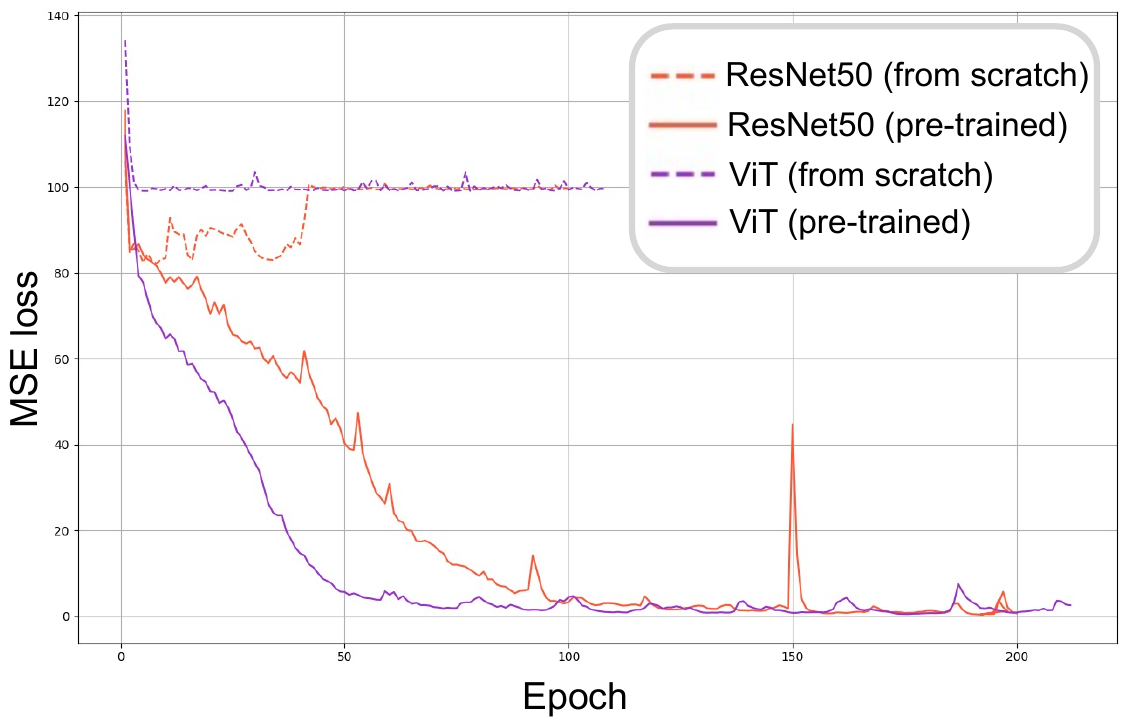} 
\caption{The regression losses of different safety prediction modules during training.}
\label{fig_abl}
\end{figure}

In this subsection, we conduct ablation studies on the safety prediction module to investigate which vision encoder can achieve better prediction performance. In Figure \ref{fig_abl} and Table \ref{tab_abl}, we compare the ResNet encoder with the ViT encoder, and their pre-trained and from-scratch versions. The $y$ axis of Figure \ref{fig_abl} is the MSE loss, and the $x$ axis is the training epoch. With pre-training on the ImageNet dataset, both the ViT encoder and the ResNet encoder can converge to a small regression loss, and the ViT encoder has a faster convergent speed. Therefore, we employ the pre-trained ViT encoder as the vision backbone in the safety prediction module of the Safe Planner framework.

We further investigate the relationship between the convergent regression losses with the large model planning performance in the Chair (Hard) task. From Table \ref{tab_abl}, we can see that a smaller regression loss of the safety module reduces the collision numbers in the robot task. As for the vision encoders from scratch, the regression loss can hardly be optimized, so the collision numbers for these models are consistent with that in the last row of Table \ref{sim_colli} without the safety module guidance.

\input{tables/abl_new.tex}

\section{Conclusion}

To address the problem of large model planners lacking physics-grounded safety awareness, we propose a novel Safe Planner framework, which guides the large pre-trained models with a learnable safety prediction module to accomplish safe and executable robot task planning. This safety prediction module trained in simulated environments can be effectively transferred to real-world tasks.  Experiment results on both simulators and real robots demonstrate that Safe Planner can significantly improve safety and avoid collisions during task execution. Beyond that, we conduct thorough ablation studies on the safety module design and find that the pretraining of the vision encoder is quite important. 

For future work, the scene understanding ability of large pre-trained models needs to be further enhanced, so that the planner's reliance on the state predicates can be alleviated. Another interesting future direction is large model quantization. With a light model, the inference speed can be improved, which can promote real-time planning for the embodied agents.

\bibliography{aaai25}

\end{document}

%% file: tables/sim_colli.tex
\begin{table*}[t!]
\centering
  \begin{tabular}{c|c|cccc}
  \toprule
    \textit{Scene} & \textit{Mode} & Safe Planner & Safe Planner w/o SM & SayCan & PROGPROMPT \\ 
    \midrule
    \multirow{2}*{Table} & Easy & \textbf{2.44} & 3.80 & 3.40 & 3.05  \\
     ~  & Hard & \textbf{4.33} & 5.42 & 6.03 & 6.23   \\
     \hline
    \multirow{2}*{Counter} & Easy & \textbf{2.07} & 2.63 & 3.08 & 3.13 \\
    ~ & Hard & \textbf{1.73} & 3.20 & 3.53 & 3.71 \\
    \hline
    \multirow{2}*{Chair} & Easy & \textbf{1.26} & 2.30 & 2.28 & 2.26 \\ 
    ~ & Hard & \textbf{6.21} & 10.59 & 10.50 & 10.51 \\
    \bottomrule
  \end{tabular}
  \caption{The comparison results of the average collision numbers.}
  \label{sim_colli}
\end{table*}

%% file: tables/sim_sr.tex
\begin{table*}[t!]
\centering
  \begin{tabular}{c|c|cccc}
  \toprule
    \textit{Scene} & \textit{Mode} & Safe Planner & Safe Planner w/o SM & SayCan & PROGPROMPT \\ 
    \midrule
    \multirow{2}*{Table} & Easy & \textbf{0.70} & 0.67 & 0.43 & 0.63  \\
     ~  & Hard & 0.73 & \textbf{0.80} & 0.67 & 0.73   \\
     \hline
    \multirow{2}*{Counter} & Easy & \textbf{0.40} & \textbf{0.40} & 0.29 & 0.38 \\
    ~ & Hard & \textbf{0.40} & \textbf{0.40} & 0.37 & 0.33 \\
    \hline
    \multirow{2}*{Chair} & Easy & 0.90 & \textbf{0.95} & 0.90 & 0.91 \\ 
    ~ & Hard & 0.82 & \textbf{0.83} & 0.81 & 0.80 \\
    \bottomrule
  \end{tabular}
  \caption{The comparison results of the average task success rates.}
  \label{sim_sr}
\end{table*}

%% file: tables/real_colli.tex
\begin{table}[h!]
\centering
\begin{tabular}{|c|c|c|c|}
\hline
 & Easy & Medium & Hard \\ \hline
Safe Planner &  \textbf{0.2}&  \textbf{1.0}&  \textbf{1.0}\\ \hline
Safe Planner w/o SM &  2.0&  4.2&  2.1\\ \hline
\end{tabular}
\caption{The average \textit{collisions} in the real robot tasks.}
\label{real_colli}
\end{table}

%% file: tables/real.tex
\begin{table}[h!]
\centering
\begin{tabular}{|c|c|c|c|}
\hline
 & Easy & Medium & Hard \\ \hline
Safe Planner &  \textbf{1.0}&  \textbf{1.0}&  \textbf{0.7}\\ \hline
Safe Planner w/o SM &  \textbf{1.0}&  \textbf{1.0}&  0\\ \hline
\end{tabular}
\caption{The average success rates in the real robot tasks.}
\label{real_sr}
\end{table}

%% file: tables/abl_new.tex
% \documentclass{article}
% \usepackage{graphicx}
% \usepackage{booktabs}

% \begin{document}
	
	\begin{table}[htbp]
		\centering
		\begin{tabular}{lcc}  
			\toprule
			\textbf{model}  & \textbf{convergent loss} & \textbf{collisions}  \\
			\midrule
   Resnet50 (from scratch)    & 100.1 & 10.59  \\
			ResNet50 (pre-trained)     & 1.9  & 6.48 \\
			ViT (from scratch)   & 99.2  & 10.52  \\
			ViT (pre-trained)   & \textbf{1.2}  & \textbf{6.21}  \\
			\bottomrule
		\end{tabular}
		\caption{The convergent regression losses of safety modules and the collisions in the Chair (Hard) task.}
		\label{tab_abl}
	\end{table}
	
% \end{document}